\documentclass[conference]{IEEEtran}

\IEEEoverridecommandlockouts
\usepackage{cite}
\usepackage{makecell} 
\usepackage{graphicx}
\usepackage{mathnotation} 
\usepackage{enumitem}
\usepackage[caption=false,font=footnotesize]{subfig}
\usepackage{tikz}
\usepackage{xspace}
\usepackage[bookmarks=false]{hyperref}

\usepackage[capitalize,noabbrev]{cleveref}
\usetikzlibrary{positioning}

\ifodd 1
\newcommand{\congr}[1]{{\color{blue}#1}}
\else
\newcommand{\congr}[1]{#}
\fi
\ifodd 1
\newcommand{\congc}[1]{{\color{red}(Cong: #1)}}
\else
\newcommand{\congc}[1]{}
\fi

\ifodd 1
\newcommand{\jing}[1]{{\color{red}#1}}
\else
\newcommand{\jing}[1]{#}
\fi
\ifodd 1
\newcommand{\jingc}[1]{{\color{red}(JY: #1)}}
\else
\newcommand{\jingc}[1]{}
\fi

\ifodd 1
\newcommand{\li}[1]{{\color[HTML]{007FFF}#1}} 
\else
\newcommand{\li}[1]{#}
\fi
\ifodd 1
\newcommand{\lic}[1]{{\color[HTML]{007FFF}(Li: #1)}} 
\else
\newcommand{\lic}[1]{}
\fi

\def \alg{\texttt{CHOOSE}\xspace}

\title{Chain-of-Thought Enhanced Shallow Transformers for Wireless Symbol Detection
}
\author{
    Li Fan\textsuperscript{1}, Peng Wang\textsuperscript{2}, Jing Yang\textsuperscript{1,2}, Cong Shen\textsuperscript{1} \\
    \textsuperscript{1}Charles L. Brown Department of Electrical and Computer Engineering, University of Virginia, USA \\
    \textsuperscript{2}Department of Computer Science, University of Virginia, USA \\
    E-mails: \texttt{\{lf2by,pw7nc,yangjing,cong\}@virginia.edu}
}

\begin{document}

\maketitle

\begin{abstract} 

Transformers have shown potential in solving wireless communication problems, particularly via in-context learning (ICL), where models adapt to new tasks through prompts without requiring model updates. However, prior ICL-based Transformer models rely on deep architectures with many layers to achieve satisfactory performance, resulting in substantial storage and computational costs. In this work, we propose \underline{CH}ain \underline{O}f th\underline{O}ught \underline{S}ymbol d\underline{E}tection (\alg), a CoT-enhanced shallow Transformer framework for wireless symbol detection. By introducing autoregressive latent reasoning steps within the hidden space, \alg\ significantly improves the reasoning capacity of shallow models (1–2 layers) without increasing model depth. This design enables lightweight Transformers to achieve detection performance comparable to much deeper models, making them well-suited for deployment on resource-constrained mobile devices. Experimental results demonstrate that our approach outperforms conventional shallow Transformers and achieves performance comparable to that of deep Transformers, while maintaining storage and computational efficiency. This represents a promising direction for implementing Transformer-based algorithms in wireless receivers with limited computational resources. 

\end{abstract}

\section{Introduction}

In-context learning (ICL) \cite{brown2020language} is a fundamental capability of Transformers \cite{vaswani2017attention}, allowing models to perform new tasks by conditioning on a sequence of input-output demonstrations without requiring parameter updates. A typical ICL prompt takes the form $(y_1, x_1, \dots, y_n, x_n, y_{n+1})$, where the model predicts $\hat{x}_{n+1}$ based on the preceding demonstration pairs $(y_i, x_i)$ and the new query $y_{n+1}$. It has been shown that ICL-based Transformers can approximate a broad class of functions under appropriate conditions \cite{garg2022can,bai2023transformers}.

Wireless communication tasks, such as symbol detection, naturally align with this ICL formulation. In block fading channels, the random channel state remains constant within a coherence block of symbols. By conditioning on a small number of pilot signal pairs, an ICL-based Transformer can implicitly infer the latent channel and accurately detect subsequent symbols. Recent studies have shown that Transformers trained under this paradigm can achieve near-optimal performance in various wireless scenarios \cite{kunde2025transformers, zecchin2024context, fan2025arXiv}. Notably, \cite{fan2025arXiv} shows that ICL-based Transformers can flexibly interpolate between coherent and non-coherent detection strategies.

Despite these advantages, most existing ICL-based Transformers rely on \emph{deep} architectures to achieve competitive performance. However, such deep models introduce significant computational and memory overhead, making them impractical for deployment on resource-constrained mobile devices \cite{xu2018designing}. Even during inference, these models suffer from high latency and energy consumption, which limits their applicability in real-time and low-power wireless systems.

To address these challenges, we draw inspiration from Chain-of-Thought (CoT) reasoning \cite{wei2022chain, wang2022self}, a technique originally developed to improve multi-step reasoning in large language models (LLMs). CoT encourages models to generate intermediate reasoning steps before making final predictions, thereby enhancing both accuracy and interpretability. Extensions of this idea, such as \cite{deng2024explicit, hao2024training}, demonstrate that CoT can operate effectively in the latent embedding space, without requiring explicit symbolic outputs.

More recently, both theoretical and empirical studies suggest that CoT reasoning can substantially improve the expressiveness of shallow Transformers, which have significantly fewer layers. For example, \cite{li2024chain, merrill2024the} provide formal analyses showing that CoT reasoning increases the functional capacity of decoder-only architectures. Additionally, \cite{huang2025transformers} demonstrates that even a single-layer Transformer can emulate multi-step gradient descent through autoregressive reasoning, highlighting the potential of shallow models to approximate complex mappings when structured appropriately.

Motivated by these insights, we propose \alg -- \underline{CH}ain \underline{O}f th\underline{O}ught \underline{S}ymbol d\underline{E}tection. \alg is a CoT-enhanced shallow Transformer architecture (1–2 layers) for in-context wireless symbol detection. To the best of our knowledge, this is the first work to integrate CoT reasoning into ICL-based Transformers for wireless applications. By introducing intermediate reasoning steps within the Transformer’s latent space, \alg significantly improves the expressiveness of shallow models without increasing their depth. Extensive experiments demonstrate that \alg\ consistently outperforms conventional shallow baselines and matches the accuracy of much deeper models, while greatly reducing storage and computation costs, making it well-suited for edge deployment. The key contributions of this work are:

\begin{itemize}[leftmargin=*]\itemsep=0pt
    \item We propose \alg, a CoT-enhanced shallow Transformer for in-context wireless symbol detection, achieving strong performance with minimal depth.
    \item We introduce a training strategy that enables latent reasoning without explicit supervision on the intermediate steps.
    \item We show through extensive experiments that CoT reasoning significantly improves both the expressiveness and detection accuracy of shallow Transformers while maintaining low computational overhead.
\end{itemize}

\vspace{-3pt}
\section{Wireless System Model}
\label{sec:system_model}

We consider a narrowband single-input single-output (SISO) communication system operating over a block fading channel, following the standard formulation in \cite{tse2005fundamentals}. The complex-valued channel coefficient $h \sim P_H$ remains constant within each coherence block of $T$ symbol periods and is sampled independently and identically (i.i.d.) across different blocks. The received signal at time $t$ is modeled as:
\begin{equation}
y_t = h x_t + z_t,
\end{equation}
where $x_t \in \mathcal{X}$ is the transmitted symbol, sampled uniformly at random from a modulation constellation. The noise term $z_t$ is modeled as additive white Gaussian noise (AWGN), drawn from the distribution $z_t \sim \mathcal{CN}(0, \sigma^2)$. 
The average transmit power is normalized to unity, i.e.,
$ \mathbb{E}[\|x_t\|^2] = 1$, and the channel is assumed to have unit variance as well, leading to the received signal-to-noise ratio (SNR) as $ \text{SNR} = \frac{1}{\sigma^2}.$

Within each coherence block of length $T$, the first $k$ symbols are designated as pilots, which are known to both the transmitter and receiver. These pilot pairs $(y_i, x_i)$ serve as contextual information to help infer the underlying channel condition. The objective is to detect the remaining $T - k$ data symbols using only the received signals and pilot references, without relying on explicit channel estimation.

\section{ICL-Based Symbol Detection}

\subsection{ICL-Based Symbol Detection Formulation}
\label{sec:ICL_formulation}
We formulate wireless symbol detection as a regression task under the ICL paradigm. In this setting, the model predicts the transmitted symbol for a new query input based on a sequence of pilot demonstration examples, without updating its parameters.

Each detection task $\tau$ is defined by a latent channel coefficient $h$ and noise level $\sigma^2$, drawn from the joint task distribution $P_\tau = P_H P_{\sigma^2}$. The receiver has no direct knowledge of the specific task realization $\tau$ and must infer the relevant channel state purely from the observed pilot pairs. Conditioned on a fixed task $\tau$, both the pilot pairs $(x_i, y_i)$ and the query $(x_t, y_t)$ for $t = k+1, \dots, T$ are  i.i.d. sampled.
We denote the pilot context as:
\begin{equation}
D_k^\tau = \{(y_1, x_1), \dots, (y_k, x_k)\},
\end{equation}
where each $(y_i, x_i)$ is a received-transmitted symbol pair under the same task $\tau$. Given a new query observation $y_t, t = k+1, \dots, T$, the model predicts the corresponding transmitted symbol via a learned function $f$:
\begin{equation}
\hat{x}_t = f(D_k^\tau, y_t).
\end{equation}

The model's performance with $k$ pilot symbols is evaluated using the mean squared error (MSE), averaged over the task distribution:
\begin{equation}
\text{MSE}_k = \mathbb{E}_{\tau \sim P_\tau} \, \mathbb{E}_{(D_k^\tau, y_t)|\tau} \left[ \| x_t - f(D_k^\tau, y_t) \|^2 \right].
\end{equation}

In addition to MSE, we also evaluate the Symbol Error Rate (SER)  with $k$ pilot symbols: 
\begin{equation}
\text{SER}_k = \mathbb{E}_{\tau \sim P_\tau} \, \mathbb{E}_{(D_k^\tau, y_t)|\tau} \left[ \mathcal{P}(\hat{x}_t) \neq x_t \right].
\end{equation}
As the Transformer has continuous-valued outputs, we add a nearest-neighbor projection operation, which is formally defined as:
\begin{equation}
\mathcal{P}(\hat{x}_t) = \arg\min_{x' \in \mathcal{X}} \| \hat{x}_t - x' \|,
\end{equation}
where $\mathcal{X}$ denotes the modulation constellation set. This projection selects the constellation symbol closest to the soft estimate $\hat{x}_t$ in Euclidean distance. 

\subsection{Vanilla In-Context Symbol Detection}
\label{sec:Vanilla_ICL}

\begin{figure}[!htbp]
    \centering
    \includegraphics[width=0.9\linewidth]{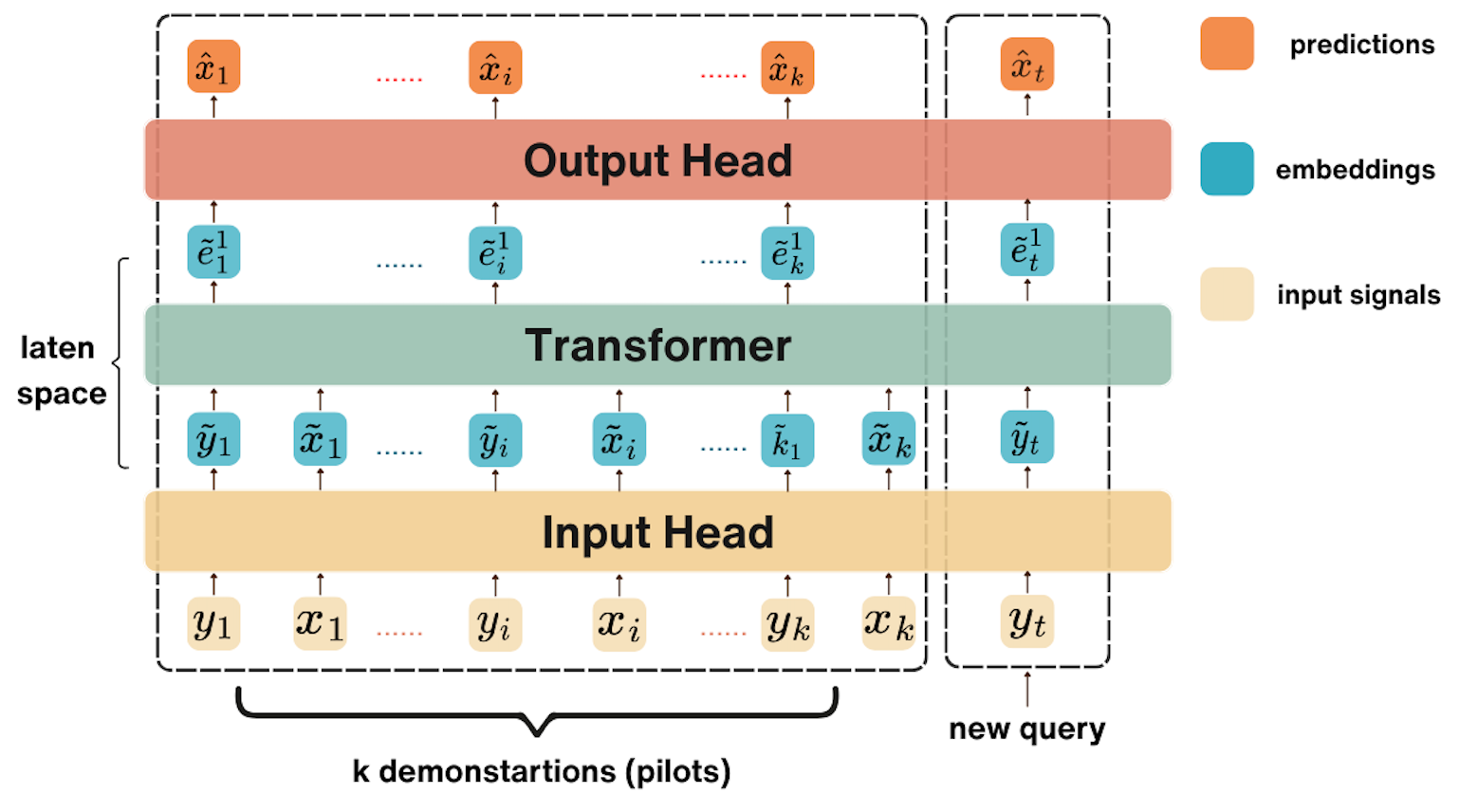}
    \vspace{-5pt}
    \caption{Vanilla in-context symbol detection architecture. The input head embeds the pilot pairs and query. A causal Transformer processes the embedded sequence under masked self-attention, and the output head predicts the query symbol from its final embedding.}
    \label{fig:vanilla-icl}
\end{figure}

We introduce the baseline architecture for ICL-based symbol detection as illustrated in \cref{fig:vanilla-icl}, following \cite{kunde2025transformers, zecchin2024context}. 
The model receives a flattened sequence of $k$ pilot demonstrations, where the $i$-th demonstration consists of a received symbol $y_i$ and its corresponding transmitted symbol $x_i$, followed by a new query symbol $y_{t}$ whose transmitted symbol $x_{t}$ is to be predicted for $t = k+1, \dots, T$. Each complex symbol is decomposed into its real (I) and imaginary (Q) components and represented as a real-valued interleaved I/Q sequence.

The input sequence is first mapped into a higher-dimensional embedding space via the \textbf{Input Head}. The resulting embeddings are then processed by a causal \textbf{Transformer} backbone—typically composed of multiple stacked decoder layers (as in GPT-2 in our experiments)—where masked self-attention ensures that each position can only attend to preceding tokens. Finally, the hidden embedding at the query position is passed through the \textbf{Output Head} to produce a continuous-valued symbol estimate $\hat{x}_{t} \in \mathbb{C}$.

\section{CoT-Enhanced In-Context Symbol Detection}

\begin{figure}[!htbp]
    \centering
    \includegraphics[width=0.95\linewidth]{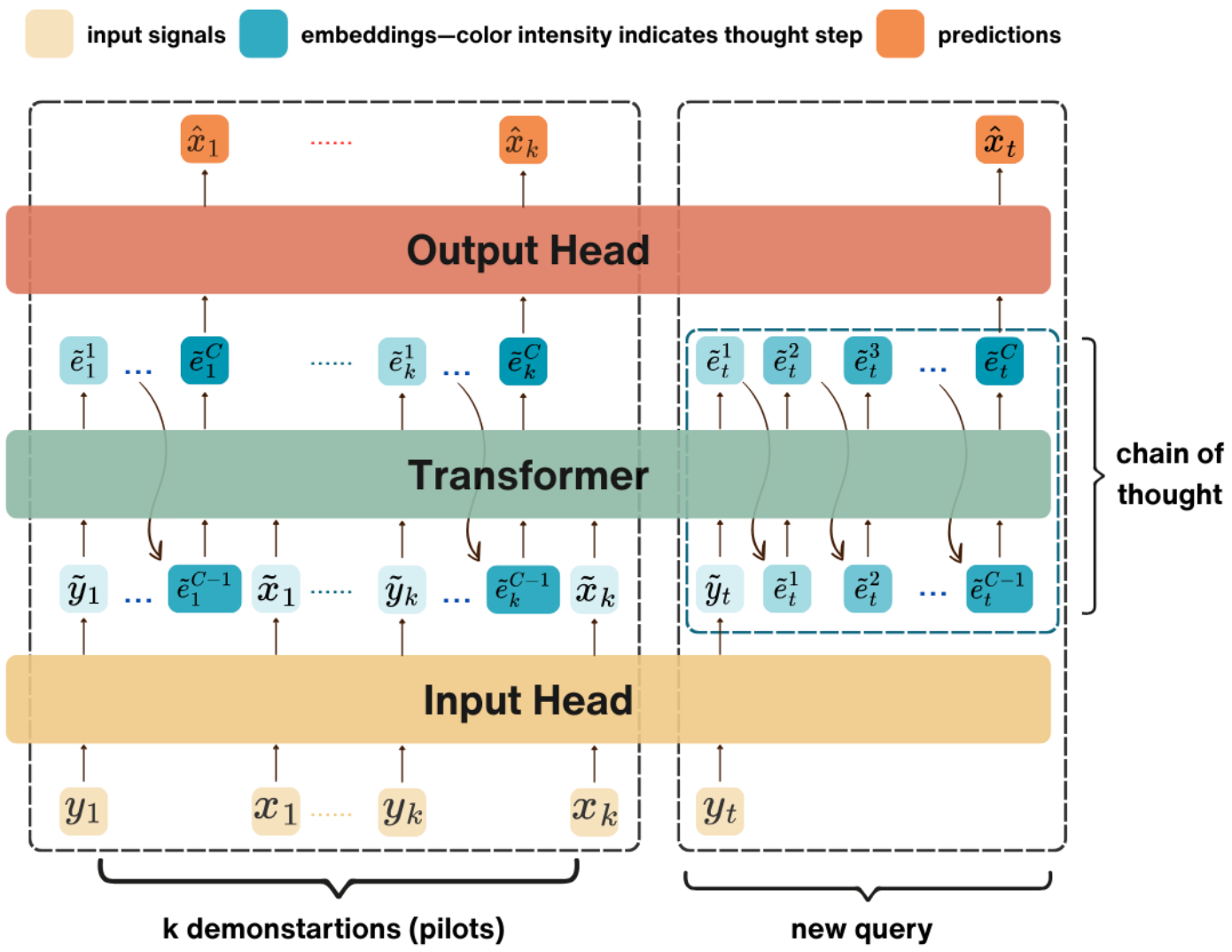}
    \vspace{-10pt}
    \caption{Architecture of \alg. The input head embeds pilot demonstrations and the query symbol into the latent space. A shallow Transformer backbone processes the sequence with causal masking. At the query position, $C$ latent thoughts embeddings $\Tilde{e}_t^{1}, \dots, \Tilde{e}_t^{C}$ are generated via autoregressive feedback within the hidden space. The final thought $\Tilde{e}_t^{C}$ is passed through the output head to produce the symbol estimate $\hat{x}_t$.}
    \label{fig:model_CoT}
\end{figure}

In order to reduce the memory and computation requirements so that Transformer-based wireless symbol detection can be realized in resource-constrained devices, we need to reduce the size of the Transformer. However, directly reducing the depth often leads to unsatisfactory inference performance. To solve this dilemma, we propose \alg, a CoT-enhanced shallow Transformer architecture that incorporates autoregressive latent reasoning steps -- referred to as \emph{thoughts} -- within the hidden space at the query position, as illustrated in \cref{fig:model_CoT}. Inspired by the latent CoT mechanism in \cite{hao2024training}, \alg\ embeds multi-step reasoning directly into the Transformer’s hidden space by iteratively refining the query embedding through a sequence of thought steps before generating the final prediction. This approach allows shallow Transformers (only 1 or 2 layers) to emulate the reasoning capacity of deeper Transformers without increasing model depth, offering a storage- and compute-efficient solution well-suited for edge-device deployment, while fully preserving the standard ICL input-output structure.

\vspace{-10pt}
\vspace{-10pt}

\subsection{Model Architecture}

The overall structure of \alg\ maintains the standard ICL framework, where the input sequence consists of pilot demonstrations followed by the query symbol. The input is first embedded through an \textbf{Input Head}, and then processed by a shallow causal Transformer with masked self-attention to respect the sequential ordering.

The key innovation lies in the query position. Instead of predicting directly from the Transformer output as in vanilla ICL, \alg\ introduces $C$ latent thought steps, where the query embedding is iteratively refined. At each reasoning step $j = 1, \dots, C$, the query embedding attends to the entire input sequence -- including pilot demonstrations and the query itself -- as well as previous thought embeddings:
\begin{equation}
\Tilde{e}_t^{j} = \text{Transformer}\big(D_{k}, y_t, \Tilde{e}_t^{1}, \dots, \Tilde{e}_t^{j-1}\big),
\end{equation}
where $D_{k}$ represents the $k$ pilot pairs and $y_t$ is the query received symbol.
After $C$ iterations of latent reasoning, only the final thought $\Tilde{e}_t^{C}$ is passed through the \textbf{Output Head} to produce the soft symbol estimate:
$ \hat{x}_t = \text{OutputHead}(\Tilde{e}_t^{C}).$

Notably, the intermediate thoughts $\Tilde{e}_t^{1}, \dots, \Tilde{e}_t^{C-1}$ are left \emph{unsupervised}, allowing the model to freely explore the latent reasoning trajectory. Supervision is applied solely on the final prediction $\hat{x}_t$, encouraging the emergence of effective multi-step reasoning without the need for explicit guidance at intermediate steps.

A key advantage of this design is its full architectural compatibility with widely adopted Transformer-based models for wireless systems. The latent CoT mechanism can be seamlessly integrated into existing ICL pipelines without changing the input or output structure, providing a simple yet powerful approach to enhance reasoning depth while maintaining a lightweight footprint suitable for edge deployment.

\subsection{Training Methodology}
\label{sec:training}
\subsubsection{Data Generation}
Training data are synthetically generated following the model described in \cref{sec:system_model}. 
To enhance robustness, each training sample is simulated with an SNR uniformly sampled from a specified range: 25–35 dB for 16QAM and 30–45 dB for 64QAM in our experiments. This exposes the model to diverse channel conditions during training. Evaluation is performed at fixed SNR values to ensure consistent benchmarking across methods.

\subsubsection{Latent CoT Training and Loss Design}

The training objective for \alg\ follows its autoregressive latent reasoning mechanism, where at each query position $t$, the model generates $C$ latent thoughts and outputs the symbol estimate $\hat{x}_t$ from the final thought $\Tilde{e}_t^{C}$. Supervision is applied only to the final prediction at each query, while intermediate thoughts remain unsupervised.

Given the full input sequence of $T$ demonstration pairs, the overall training loss is defined as the sum of the MSE loss across all query positions:
\begin{equation}
    \mathcal{L}_{\text{MSE}} = \sum_{t=1}^{T} \mathbb{E} \left[ \|x_t - \hat{x}_t\|^2 \right],
\end{equation}
where $\hat{x}_t$ is obtained via:
\begin{equation}
    \hat{x}_t = \text{OutputHead}(\Tilde{e}_t^{C}).
\end{equation}
No explicit loss is applied to the intermediate thoughts $\Tilde{e}_t^{1}, \dots, \Tilde{e}_t^{C-1}$, allowing the model to freely explore its latent reasoning trajectory without external constraints. This design encourages the emergence of effective multi-step reasoning patterns, enabling shallow architectures to achieve high detection accuracy with minimal model depth.

\subsection{Parameter Efficiency and Model Compactness}
\label{sec:parameter}
The \alg\ architecture is designed to be highly parameter-efficient, adopting a shallow Transformer backbone with only 1 or 2 layers, an embedding dimension of $d = 32$, and 4 attention heads. The number of latent thoughts $C$ is set between 1 and 4, resulting in a total parameter count of approximately {$27,000$}.

In contrast, the Transformer baseline from \cite{fan2025arXiv} utilizes $L = 8$ layers, $d = 64$ embedding dimension, and $h = 8$ attention heads, yielding around 0.42 million parameters. Despite this nearly \textbf{tenfold reduction in model size}, \alg\ achieves comparable detection performance, which will be evident in \Cref{sec:exp}, by leveraging latent CoT reasoning to enhance the expressiveness of shallow architectures.
This compact model footprint directly translates to lower storage requirements and reduced memory consumption, providing a practical solution for edge-device deployment in wireless communication systems where hardware resources are often highly constrained.

\subsection{Computation Efficiency via KV Caching}
\label{sec:computation}
Another important feature of \alg\ is that although it introduces $C$ autoregressive latent reasoning steps per query, the associated computational overhead remains minimal due to the use of key-value (KV) caching \cite{jin2017netcache} -- a standard optimization in Transformer architectures such as GPT-2.

The latent CoT reasoning loop in \alg\ is fully compatible with causal masked self-attention. At each thought step $j$, the query embedding $\Tilde{e}_t^{j}$ attends to the entire input sequence (including the pilot demonstrations and query) as well as its own previous thought embeddings $\Tilde{e}_t^{1}, \dots, \Tilde{e}_t^{j-1}$. Crucially, because the input tokens and earlier thoughts remain unchanged across reasoning steps, their key and value projections can be precomputed, stored, and reused. At each new step, the model only needs to compute the projections for the latest thought and perform attention against the cached keys and values.
This enables the multi-step latent CoT process to operate with efficiency comparable to a single forward pass over an extended sequence.

Combined with the shallow Transformer backbone (1 or 2 layers) and moderate chain length ($C = 1$ to $4$), KV caching enables \alg\ to support expressive multi-step reasoning while maintaining low computational cost during inference. This synergy between latent CoT reasoning and caching ensures that \alg\ remains well-suited for deployment on resource-constrained wireless edge devices.

\section{Experiments}
\label{sec:exp}

\subsection{Experimental Setup}

We evaluate \alg\ on a SISO symbol detection task under the Rayleigh block fading channel model described in \cref{sec:system_model}. The transmitted symbols are uniformly drawn from a given modulation constellation (16QAM or 64QAM), with each coherence block containing $T = 11$ symbols. Model training follows the synthetic data generation procedure outlined in \cref{sec:training}. During evaluation, we test across fixed SNR levels, varying both the number of pilot symbols $k$ and the number of latent reasoning steps $C \in \{1, 2, 3,4\}$ for \alg.
Performance is assessed using MSE and SER, as defined in \cref{sec:ICL_formulation}. The continuous symbol prediction $\hat{x}_t$ is projected to the nearest constellation point using nearest-neighbor mapping for SER computation. All results are averaged over 10,0000 independently generated test tasks for each configuration to ensure statistical reliability.

\subsection{Baselines Algorithms}

We compare \alg\ against representative baselines. To ensure consistency, all models use an embedding dimension of 32 and 4 attention heads. Only the number of Transformer layers is varied to assess the effect of depth in both vanilla ICL and \alg\ models.

\paragraph{Vanilla ICL}
This baseline follows the standard ICL architecture described in \cref{sec:Vanilla_ICL}, where the Transformer directly maps the input context (pilot demonstrations and query) to the symbol estimate, without latent CoT reasoning. To investigate the effect of model capacity, we evaluate Vanilla ICL Transformers with varying depths: 1, 2, and 4 layers. These models share the same input-output structure as \alg\ and serve as non-CoT baselines for fair comparison.

\paragraph{MMSE Estimator}

We implement the minimum mean squared error (MMSE) estimator\cite{tse2005fundamentals} as an oracle baseline, providing the theoretical lower bound on MSE under the Rayleigh block fading model. This estimator assumes knowledge of the channel distribution $h \sim \mathcal{CN}(0,1)$ and AWGN noise $z_t \sim \mathcal{CN}(0, \sigma^2)$ but not the instantaneous channel realization.
Given the pilots $x_{1:k}$, received sequence $y_{1:k}$, and query $y_{k+1}$, the MMSE estimate corresponds to the conditional mean estimator (CME). The posterior probability of the transmitted symbol $x_{k+1} = x$ for each $x \in \mathcal{X}$ is proportional to the Gaussian likelihood:
\begin{equation}
\begin{split}
    &\mathbb{P}(x_{k+1} = x \mid y_{1:k+1}, x_{1:k})
    \propto\  \\
    &\frac{1}{\sqrt{|\det(C(x))|}}  \exp\left( -\frac{1}{2} y_{1:k+1}^\mathrm{H} C^{-1}(x) y_{1:k+1} \right).
\end{split}
\end{equation}
where the covariance matrix $C(x)$ is given by:
\begin{equation}
    C(x) = \frac{1}{2} X_{1:k+1} X_{1:k+1}^\mathrm{H} + \sigma^2 I_{k+1},
\end{equation}
and $X_{1:k+1}$ includes both the $k$ pilot symbols and the candidate $x$ at position $k+1$.
The MMSE estimate is then obtained as the weighted sum over the constellation:
\begin{equation}
    \hat{x}_{k+1}^{\mathrm{MMSE}} = \sum_{x \in \mathcal{X}} x \cdot \mathbb{P}(x_{k+1} = x \mid y_{1:k+1}, x_{1:k}).
\end{equation}

This oracle provides the optimal continuous prediction under MSE loss and serves as a reference for evaluating learning-based models like \alg.

\subsection{Results and Analysis}

\subsubsection{Symbol Detection Performance}

\begin{figure}[!htbp]
    \centering
    \subfloat[16QAM at 30 dB SNR.]{
        \includegraphics[width= 0.9\linewidth]{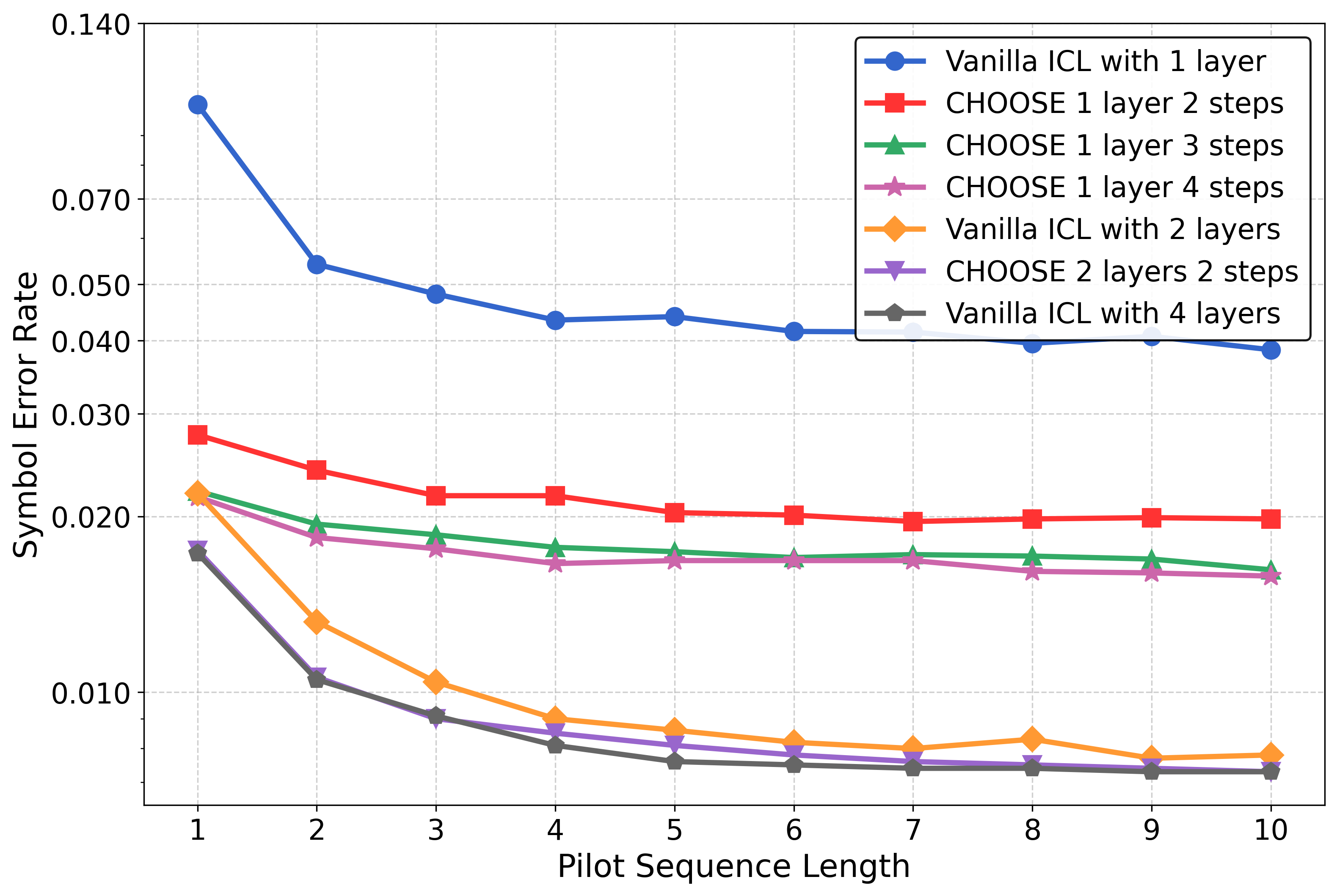}
        \label{fig:ser-16qam}
    }\\[1ex]
    \subfloat[64QAM at 40 dB SNR.]{
        \includegraphics[width= 0.9\linewidth]{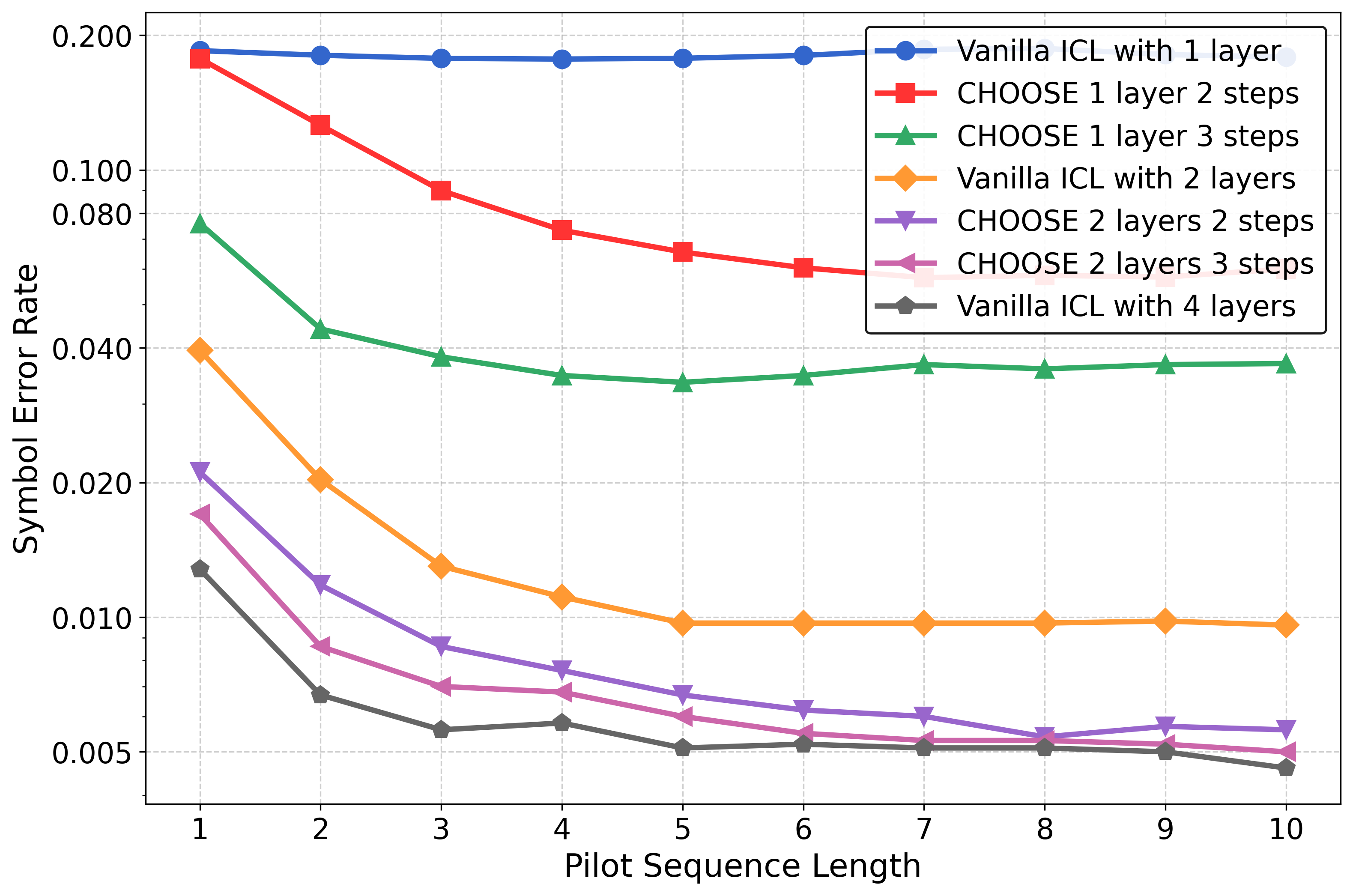}
        \label{fig:ser-64qam}
    }
    \caption{Symbol error rate vs. pilot sequence length for \alg and vanilla ICL models. \alg models achieve strong performance with fewer layers and reasoning steps.}
    \label{fig:ser-combined}
\end{figure}

\Cref{fig:ser-16qam,fig:ser-64qam} report the SER of \alg\ on 16QAM (30 dB SNR) and 64QAM (40 dB SNR) under varying pilot lengths. We compare shallow \alg\ models (1–2 layers) with latent reasoning steps $C = 2,3,4$ against vanilla ICL baselines with 1, 2, and 4 Transformer layers. All models share the same ICL setup and differ only in architectural depth and use of latent CoT reasoning.

Results indicate that latent reasoning substantially improves performance, particularly for shallow models. On 16QAM (\cref{fig:ser-16qam}), a 1-layer \alg\ model with 4 reasoning steps achieves over $50\%$ lower SER compared to its vanilla counterpart. Furthermore, a 2-layer \alg\ model with 2 steps matches the performance of a 4-layer vanilla Transformer, demonstrating that latent reasoning effectively enhances model expressiveness without increasing depth.

On the more challenging 64QAM task (\cref{fig:ser-64qam}), this trend persists and strengthens. A 2-layer \alg\ model with 3 steps closely approaches the 4-layer vanilla model's performance, while consistently outperforming the 2-layer baseline. These findings confirm that latent CoT reasoning scales well to higher-order modulations and enables compact models to maintain high detection accuracy.

\subsubsection{Thought Refinement Dynamics}

To gain deeper insights into how \alg\ achieves strong performance with minimal architectural depth, we examine the internal progression of its latent reasoning process. Specifically, we analyze a 1-layer \alg\ model with $C = 4$ latent steps, tested on the 16QAM detection task at 30\,dB SNR. This analysis aims to visualize how the model refines its predictions across the sequence of latent thoughts during inference.

As described in \cref{sec:training}, supervision is applied exclusively to the final output—intermediate thoughts receive no direct loss feedback. To examine their behavior, we unfold the latent reasoning trajectory at inference time by passing each intermediate embedding $\Tilde{e}_t^1, \dots, \Tilde{e}_t^4$ at every query position through the shared output head. The resulting soft symbol predictions at each step are then evaluated using both MSE and SER metrics.

\begin{figure}[!htbp]
    \centering
    \subfloat[MSE of intermediate thoughts vs. MMSE oracle.]{
        \includegraphics[width=0.9\linewidth]{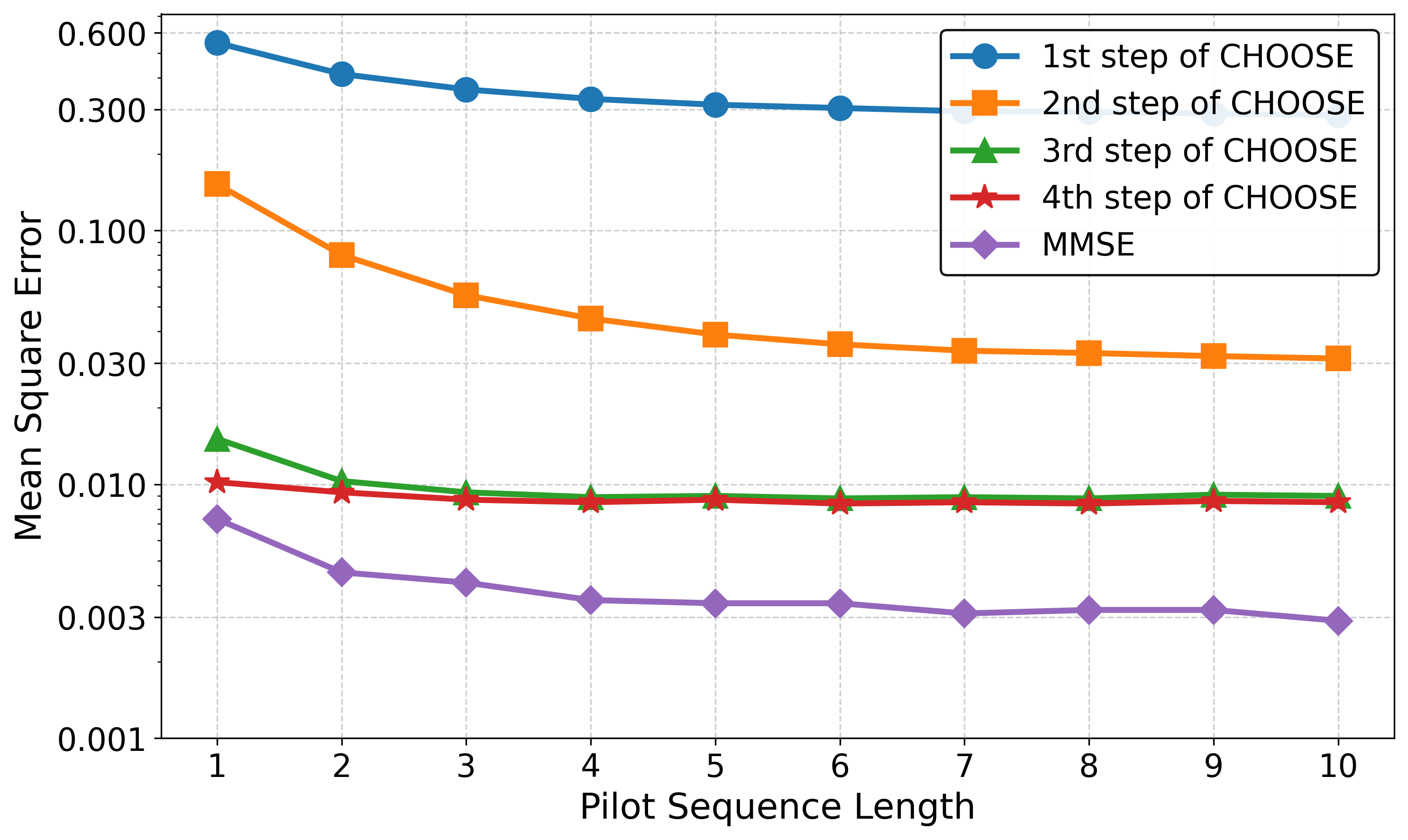}
        \label{fig:unfold-chain-mse}
    }\\[1ex]
    \subfloat[SER of intermediate thoughts.]{
        \includegraphics[width=0.9\linewidth]{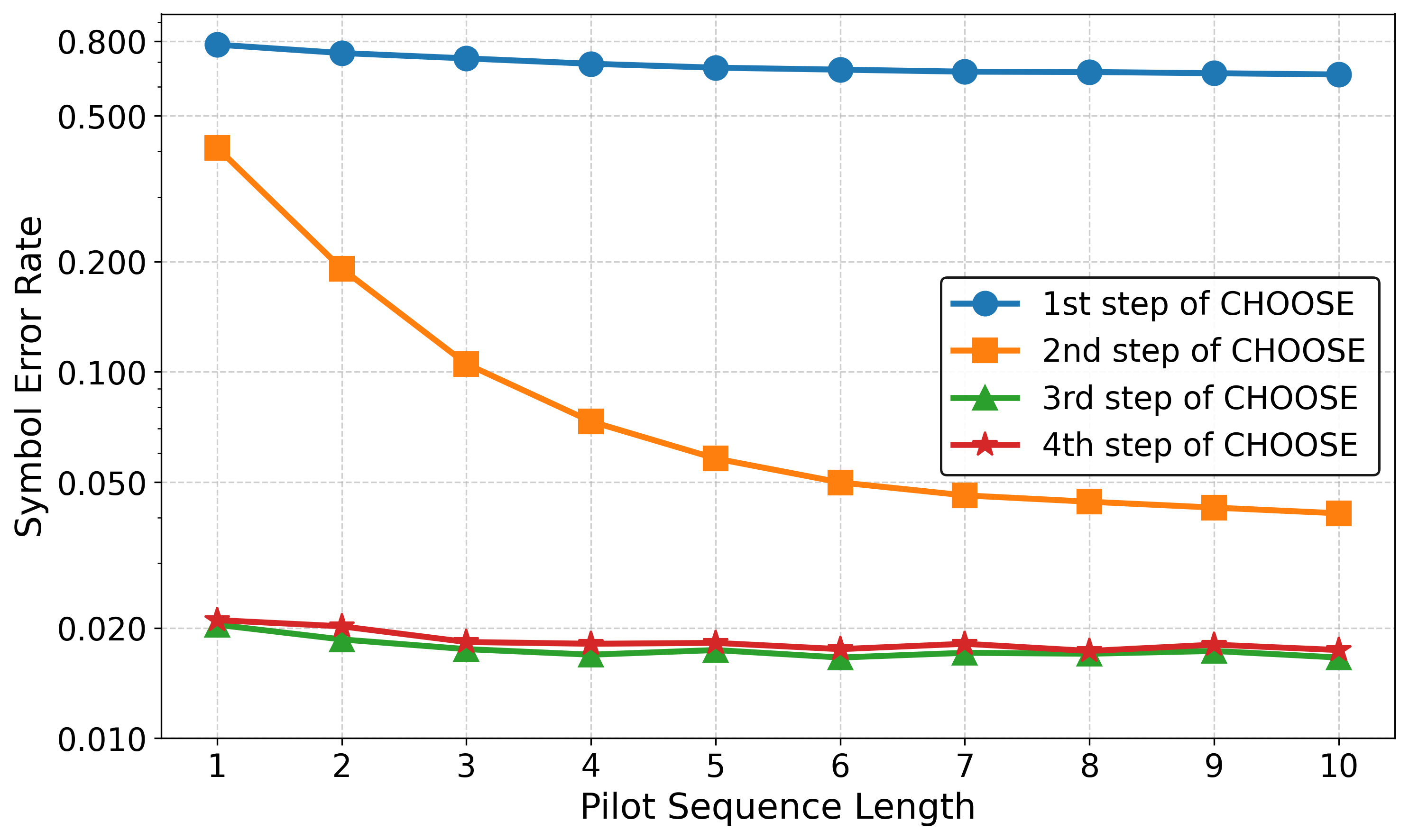}
        \label{fig:unfold-chain-ser}
    }
    \caption{Unfolded performance across different CoT steps for a 1-layer \alg\ model with $C=4$ on 16QAM at 30\,dB SNR. Each curve corresponds to a different latent thought evaluated at every query position. Despite supervision only on the final step, the model shows consistent internal refinement.}
    \label{fig:unfold-chain}
    \vspace{-10pt}
\end{figure}

As shown in \cref{fig:unfold-chain}, both MSE and SER decrease steadily across the CoT steps. The first latent thought provides a coarse approximation with relatively high error. Subsequent steps increasingly correct earlier inaccuracies, with the most substantial improvement occurring between the first and second steps. Beyond that, performance gains diminish—suggesting that the representational capacity of the single-layer backbone has been nearly saturated. Notably, the final step MSE approaches the MMSE oracle in \cref{fig:unfold-chain-mse}, though a small performance gap remains due to the architectural limitation.

This unfolding analysis illustrates that \alg\ autonomously learns to organize its latent space into an effective multi-stage refinement process. Even without intermediate supervision, the model discovers a reasoning trajectory that progressively improves prediction accuracy—driven purely by loss feedback at the final step. These results underscore the strength of latent CoT in enabling structured internal computation within compact Transformer models.

\subsubsection{Parameter and Computation Efficiency}

\begin{figure}[!htbp]
    \centering
    \includegraphics[width=0.9\linewidth]{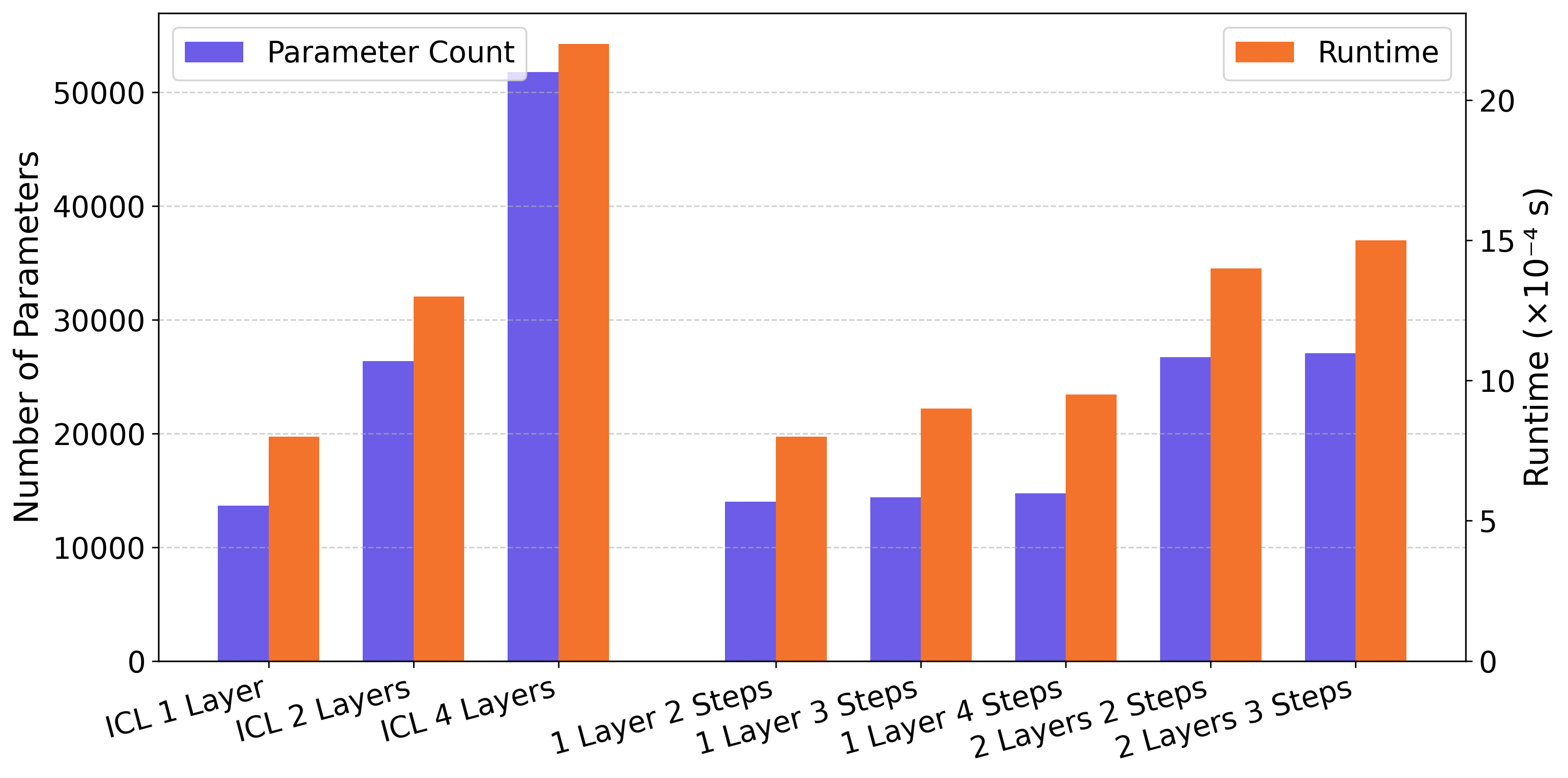}
    \vspace{-10pt}
    \caption{Comparison of parameter count and inference time across different model configurations.}
    \label{fig:parameter-comparison}
\end{figure}

To validate the lightweight nature of \alg, we compare its amount of parameters and inference efficiency with vanilla ICL baselines under various configurations. As shown in \cref{fig:parameter-comparison}, vanilla ICL models improve performance primarily by increasing Transformer depth -- e.g., the 4-layer ICL model exceeds 50,000 parameters. In contrast, \alg\ achieves comparable detection accuracy with only 1 or 2 Transformer layers and a small number of latent reasoning steps ($C = 2,3,4$), leading to significantly fewer parameters.
The slight increase in parameter amount with larger $C$ in \alg\ stems from the extended input sequence length caused by the inclusion of latent thought embeddings. This adds minimal overhead to the positional encoding and attention projections, and is still substantially more efficient than increasing the model depth.

We further evaluate computational efficiency by measuring the inference time required to process the entire test set under identical hardware conditions. As shown in \cref{fig:parameter-comparison}, \alg\ introduces only a modest increase in inference time compared to vanilla ICL at the same number of layers. This is primarily due to the autoregressive reasoning steps. However, by applying the \textit{key-value (KV) caching} technique described in \cref{sec:computation}, we effectively eliminate redundant computations during CoT-style autoregressive decoding. As a result, \alg\ achieves a favorable trade-off between computation and accuracy: a slight increase in runtime yields a substantial performance gain.

\subsubsection{Comprehensive Analysis}

Overall, our results show that \alg\ achieves high detection accuracy with significantly fewer parameters than deeper vanilla ICL models. By introducing a small number of autoregressive latent reasoning steps, \alg\ enables shallow Transformers to perform effective iterative inference, refining predictions across steps.
Despite the added reasoning steps, \alg\ remains computationally efficient. With key-value caching, inference-time overhead is minimal compared to deep ICL models, while delivering superior accuracy.
These results highlight \alg's efficiency in both model size and runtime, making it well suited for deployment in resource-constrained wireless systems.

\section{Conclusion}

We introduced \alg, a lightweight Transformer architecture for in-context wireless symbol detection that augments shallow models with autoregressive latent reasoning steps. By refining predictions through a sequence of internal thought embeddings, \alg\ achieves substantial improvements in accuracy and expressiveness without increasing model depth. Despite its shallow backbone, \alg\ delivers performance comparable to much deeper vanilla ICL baselines in both SER and MSE, while maintaining a significantly lower parameter count and efficient inference—particularly when combined with key-value caching. These results demonstrate the effectiveness of integrating latent Chain-of-Thought mechanisms into compact Transformer architectures for wireless receivers. \alg\ offers a scalable and resource-efficient solution for edge deployment, and paves the way for reasoning-based inference in physical-layer communication tasks.

\bibliographystyle{IEEEtran}
\bibliography{ref,Shen,wireless}

\begin{thebibliography}{10}
\providecommand{\url}[1]{#1}
\csname url@samestyle\endcsname
\providecommand{\newblock}{\relax}
\providecommand{\bibinfo}[2]{#2}
\providecommand{\BIBentrySTDinterwordspacing}{\spaceskip=0pt\relax}
\providecommand{\BIBentryALTinterwordstretchfactor}{4}
\providecommand{\BIBentryALTinterwordspacing}{\spaceskip=\fontdimen2\font plus
\BIBentryALTinterwordstretchfactor\fontdimen3\font minus \fontdimen4\font\relax}
\providecommand{\BIBforeignlanguage}[2]{{%
\expandafter\ifx\csname l@#1\endcsname\relax
\typeout{** WARNING: IEEEtran.bst: No hyphenation pattern has been}%
\typeout{** loaded for the language `#1'. Using the pattern for}%
\typeout{** the default language instead.}%
\else
\language=\csname l@#1\endcsname
\fi
#2}}
\providecommand{\BIBdecl}{\relax}
\BIBdecl

\bibitem{brown2020language}
T.~Brown, B.~Mann, N.~Ryder, M.~Subbiah, J.~D. Kaplan, P.~Dhariwal, A.~Neelakantan, P.~Shyam, G.~Sastry, A.~Askell \emph{et~al.}, ``Language models are few-shot learners,'' \emph{Advances in neural information processing systems}, vol.~33, pp. 1877--1901, 2020.

\bibitem{vaswani2017attention}
A.~Vaswani, N.~Shazeer, N.~Parmar, J.~Uszkoreit, L.~Jones, A.~N. Gomez, {\L}.~Kaiser, and I.~Polosukhin, ``Attention is all you need,'' \emph{Advances in neural information processing systems}, vol.~30, 2017.

\bibitem{garg2022can}
S.~Garg, D.~Tsipras, P.~S. Liang, and G.~Valiant, ``What can transformers learn in-context? a case study of simple function classes,'' \emph{Advances in Neural Information Processing Systems}, vol.~35, 2022.

\bibitem{bai2023transformers}
Y.~Bai, F.~Chen, H.~Wang, C.~Xiong, and S.~Mei, ``Transformers as statisticians: Provable in-context learning with in-context algorithm selection,'' \emph{Advances in neural information processing systems}, vol.~36, pp. 57\,125--57\,211, 2023.

\bibitem{kunde2025transformers}
V.~T. Kunde, V.~Rajagopalan, C.~S.~K. Valmeekam, K.~Narayanan, J.-F. Chamberland, D.~Kalathil, and S.~Shakkottai, ``Transformers are provably optimal in-context estimators for wireless communications,'' in \emph{Proceedings of the 28th International Conference on Artificial Intelligence and Statistics}, 2025.

\bibitem{zecchin2024context}
M.~Zecchin, K.~Yu, and O.~Simeone, ``In-context learning for mimo equalization using transformer-based sequence models,'' in \emph{IEEE International Conference on Communications Workshops}, 2024.

\bibitem{fan2025arXiv}
L.~Fan, J.~Yang, and C.~Shen, ``Transformer-based wireless symbol detection over fading channels,'' \emph{arXiv preprint arXiv:2503.16594}, 2025.

\bibitem{xu2018designing}
J.~Xu, L.~Chen, K.~Liu, and C.~Shen, ``Designing security-aware incentives for computation offloading via device-to-device communication,'' \emph{IEEE Transactions on Wireless Communications}, vol.~17, no.~9, pp. 6053--6066, 2018.

\bibitem{wei2022chain}
J.~Wei, X.~Wang, D.~Schuurmans, M.~Bosma, F.~Xia, E.~Chi, Q.~V. Le, D.~Zhou \emph{et~al.}, ``Chain-of-thought prompting elicits reasoning in large language models,'' \emph{Advances in neural information processing systems}, vol.~35, pp. 24\,824--24\,837, 2022.

\bibitem{wang2022self}
X.~Wang, J.~Wei, D.~Schuurmans, Q.~Le, E.~Chi, S.~Narang, A.~Chowdhery, and D.~Zhou, ``Self-consistency improves chain of thought reasoning in language models,'' \emph{arXiv preprint arXiv:2203.11171}, 2022.

\bibitem{deng2024explicit}
Y.~Deng, Y.~Choi, and S.~Shieber, ``From explicit cot to implicit cot: Learning to internalize cot step by step,'' \emph{arXiv preprint arXiv:2405.14838}, 2024.

\bibitem{hao2024training}
S.~Hao, S.~Sukhbaatar, D.~Su, X.~Li, Z.~Hu, J.~Weston, and Y.~Tian, ``Training large language models to reason in a continuous latent space,'' \emph{arXiv preprint arXiv:2412.06769}, 2024.

\bibitem{li2024chain}
Z.~Li, H.~Liu, D.~Zhou, and T.~Ma, ``Chain of thought empowers transformers to solve inherently serial problems,'' in \emph{The Twelfth International Conference on Learning Representations}, 2024.

\bibitem{merrill2024the}
W.~Merrill and A.~Sabharwal, ``The expressive power of transformers with chain of thought,'' in \emph{The Twelfth International Conference on Learning Representations}, 2024.

\bibitem{huang2025transformers}
J.~Huang, Z.~Wang, and J.~D. Lee, ``Transformers learn to implement multi-step gradient descent with chain of thought,'' in \emph{The Thirteenth International Conference on Learning Representations}, 2025.

\bibitem{tse2005fundamentals}
D.~Tse and P.~Viswanath, \emph{Fundamentals of wireless communication}.\hskip 1em plus 0.5em minus 0.4em\relax Cambridge university press, 2005.

\bibitem{jin2017netcache}
X.~Jin, X.~Li, H.~Zhang, R.~Soul{\'e}, J.~Lee, N.~Foster, C.~Kim, and I.~Stoica, ``Netcache: Balancing key-value stores with fast in-network caching,'' in \emph{Proceedings of the 26th symposium on operating systems principles}, 2017, pp. 121--136.

\end{thebibliography}

\end{document}